\documentclass[11pt,a4paper]{article}
\usepackage[hyperref]{acl2021}
\usepackage{times}
\usepackage{latexsym}
\usepackage{graphicx}
\usepackage{amsmath,amssymb,mathtools}

\DeclareUnicodeCharacter{2248}{\(\approx\)}
\DeclareUnicodeCharacter{2009}{\,}
\DeclareUnicodeCharacter{2013}{--}
\DeclareUnicodeCharacter{2014}{---}
\DeclareUnicodeCharacter{2018}{\textquotesingle}
\DeclareUnicodeCharacter{2019}{\textquotesingle}
\DeclareUnicodeCharacter{201C}{``}
\DeclareUnicodeCharacter{201D}{''}
\DeclareUnicodeCharacter{2193}{$\downarrow$}

\usepackage{microtype}

\aclfinalcopy 
\def\aclpaperid{eJRsxDUdxG} 

\title{Can Constructions ``SCAN" Compositionality ?}

\author{
  Ganesh Katrapati \and Manish Shrivastava\\
  International Institute of Information Technology Hyderabad\\
  \texttt{ganesh.katrapati@research.iiit.ac.in} \quad
  \texttt{m.shrivastava@iiit.ac.in}
}

\date{}

\begin{document}
\maketitle
\begin{abstract}
Sequence to Sequence models struggle at compositionality and systematic generalisation even while they excel at many other tasks. We attribute this limitation to their failure to internalise \emph{constructions}—conventionalised form–meaning pairings that license productive recombination. Building on these insights, we introduce an unsupervised procedure for mining \emph{pseudo-constructions}: variable-slot templates automatically extracted from training data. When applied to the \textsc{SCAN} dataset, our method yields large gains out-of-distribution splits: accuracy rises to \textbf{47.8\%} on \textsc{Add Jump} and to \textbf{20.3\%} on \textsc{Around Right} without any architectural changes or additional supervision. The model also attains competitive performance with \mbox{$\le 40\%$} of the original training data, demonstrating strong data efficiency. Our findings highlight the promise of construction-aware preprocessing as an alternative to heavy architectural or training-regime interventions.

\end{abstract}

\section{Introduction}

Compositionality is the principle that the meaning of a complex expression is determined by the meanings of its parts and the rules used to combine them \citep{fodor1988connectionism, marcus2003algebraic, partee1990compositionality}. It enables systematic generalisation: the ability to understand and produce novel combinations of familiar elements, a hallmark of human language competence. 

Despite the impressive empirical performance of sequence to sequence models such as RNNs, LSTMs, and Transformers, studies have consistently found that they struggle with tasks requiring compositional generalisation \citep{lake-baroni-2018-scan, hupkes2020pcfg, keysers2020cfq}. When faced with inputs that combine known primitives in unseen ways, these models frequently fail to extrapolate correctly.

Cognitive and Construction Grammar treat \emph{constructions} as form–meaning pairs composed of conventionalised components that combine with lexical items \citep{goldberg1995constructions, langacker1987foundations, croft2001radical}. For successful communication, speakers must have access to these conventionalised constructions shared within their linguistic community. The degree of conventionalisation varies across construction types: for example, idiomatic expressions like ``kick the bucket'' are fully fossilised and resist internal modification, whereas partially filled constructions such as ``the Xer the Yer'' contain open slots that can be flexibly filled to produce complete surface forms \citep{fillmore1988regularity, goldberg2006constructions}.

Inspired by this notion, we propose that modelling constructions is essential to solving the problem of compositionality. We choose the SCAN dataset - a canonical testbed for evaluating compositionality in neural models - to demonstrate our approach. We introduce a simple yet effective method of mining \emph{pseudo-constructions} and show that models trained on segmented data achieve significant improvements over standard baselines on SCAN's ADD JUMP and AROUND RIGHT splits. 

Furthermore, we demonstrate strong data efficiency: by leveraging the compositional structure, our method requires substantially less data to achieve competitive performance, especially on simpler splits. Our results suggest that carefully exposing compositional patterns during training can yield robust improvements without resorting to complex interventions.


\section{Related Work}

There have been a number of benchmarks and tasks to evaluate whether modern NLP methods including deep neural networks such as RNNs \citep{elman1990finding}, LSTMs \citep{hochreiter1997long} and Transformers \citep{vaswani2017attention} exhibit compositional behaviour. \emph{SCAN} \citep{lake-baroni-2018-scan}, \emph{COGS} \citep{kim-linzen-2020-cogs} , \emph{CFQ} \citep{keysers2020cfq}, \emph{PCFG} \citep{hupkes2020pcfg} and similar benchmarks focus on sequence prediction tasks where input sequence must be processed in a compositional manner to yield the correct sequence on the target side. 

They showed that the models do \emph{not} generalise systematically: when confronted with new combinations of words or phrases that were absent from the training data, their performance breaks down. Subsequent studies on a variety of datasets \citep{li-etal-2021-cognition, sinha-etal-2019-clutrr, liska2018memorize}, have reported similar findings. Informed by these limitations, recent work has led to multiple methods to improve compositional generalisation abilities of neural network models. 

Multiple studies have focused on disentangling syntax and semantics - \citep{russin-etal-2019-syntactic} introduced a dedicated syntactic
channel boosts SCAN accuracy dramatically , separating primitive–function pathways pushes performance to near-perfect levels \citep{li-etal-2019-cgps,jiang-bansal-2021-cgps}. Rather than separating syntax and semantics, some studies have focused on syntactic guidance. \citep{hupkes-etal-2019-attentive,baan-etal-2019-inspecting, kim-etal-2021-structural,zanzotto-etal-2020-kermit}. 

\emph{Data-centric} approaches improve compositionality by augmenting the training corpus with systematically recombined examples: GECA \citep{andreas-2020-geca}, automatically mined lexical symmetries (\textsc{LexSym}; \citealp{akyurek-andreas-2022-lexsym}), and grammar-based generators such as \textsc{CSL} \citep{qiu-etal-2022-csl} all substantially cut error rates on SCAN, COGS, and CLEVR. \citet{herzig-etal-2021-ir} insert a reversible or lossy intermediate representation between the input and the target program, doubling accuracy on CFQ MCD splits and adding 15–20 points on text-to-SQL.

Treating compositionality as a transferable skill, Meta learning approaches \citep{zhu-etal-2021-duel, lake-2019-meta,lake-baroni-2023-metaicl} push transformers beyond 70~\% accuracy on the hardest SCAN and COGS splits.

Apart from this, several studies have proposed significant modifications to the neural network architecture \citep{csordas-etal-2022-ndr, huang-etal-2024-cat} and neural–symbolic designs such as NMN, MAC, NLM, LANE, program-synthesis grammars, and the Neural-Symbolic Recursive Machine \citep{andreas-etal-2017-nmn,hudson-manning-2018-mac,dong-etal-2019-nlm,liu-etal-2022-lane,nye-etal-2020-program,li-etal-2022-nsr} which
achieve (near-)perfect compositional generalisation on datasets like SCAN, COGS and CFQ. 

While many of these approaches achieve near-perfect accuracy in datasets like SCAN and COGS, they either require data augmentation, which likely translates into training bigger models for a longer time, or they propose drastic architectural changes which have not been proven to scale beyond these benchmarks. Our method does not employ data augmentation or complex architectural changes. Our aim is show that taking insights from Cognitive Grammar and the notion of \emph{Constructions} leads to building models more capable of compositional generalisation. 

Recent work on integrating Construction Grammar (CxG) with neural models has been encouraging: fine-tuning BERT on construction-annotated corpora sharpens its encoding of construction identity and slot fillers \citep{tayyar-madabushi-etal-2020-cxgbert}, a Mandarin CxLM leverages more than ten-thousand schemata to boost cloze accuracy \citep{tseng-etal-2022-cxlm}. Yet no study has directly shown that construction-aware training itself improves systematic compositional generalisation on classic out-of-distribution tests and bridging that gap remains a challenge.

\section{Data}
Introduced by \citep{lake-baroni-2018-scan}, \emph{SCAN} contains pairs of simple navigation commands with action sequences; primitives like ``jump" map to ``\texttt{I\_JUMP}", while modifiers such as ``left", ``right", ``opposite", and ``around" compose these primitives into longer actions.

The original paper showed that models excel on a random split yet falter on novel combinations.  In the ADD JUMP split, models see the primitive ``jump" during training but must execute composed forms (e.g., ``jump twice") at test time.  \citet{loula-etal-2018-rearranging} extended this with the AROUND RIGHT split: training includes ``walk left", ``walk right", ``jump around left", and so on, while testing requires generalising to ``jump around right", forcing the model to learn that “around” modifies directions and that ``left" and ``right" are symmetric.

We focus on improving the accuracy for both these splits. 

\section{Approach}

\paragraph{Definition (Pseudo-construction).}
A \emph{pseudo-construction} is a partially specified template induced from training data, containing fixed words alongside one or more \emph{slots} represented by placeholders (e.g., \texttt{\_} or \texttt{W\_n}). Unlike fully conventionalised constructions, pseudo-constructions are derived automatically and capture recurring structural patterns that can generalise to novel inputs when the slots are filled with appropriate lexical items.

\subsection{Mining Pseudo-constructions}
A SCAN train or test set consists of both a source file, which consists of commands (``jump") and a target file which consists of actions (``\texttt{I\_JUMP}"). Given a SCAN split, we take the source file of the training set, and follow a series of steps to obtain partially filled pseudo-constructions. 

\begin{itemize}

    \item \textbf{Extracting Candidates: } For every sentence in the train source file, we extract spans of up to length of 4 tokens and add them to the candidate list. We also generate masked spans in which one or more non-consecutive words are replaced by the the token ``\texttt{\_}", effectively forming a \emph{slot} in a partially filled pseudo-construction. The candidates are then ranked according to their probabilities. 
    
    \item \textbf{Beam Decoding: } We use beam search to segment an input sentence into the best scoring sequence of pseudo-constructions and words. Test source files are not used for mining pseudo-constructions. They are segmented only using the ones induced from the training set. 

    \item \textbf{Encouraging Alignment with Target}: Partially filled pseudo-constructions like `\texttt{\_} around \texttt{\_} twice" are advantageous because the same template applies for any fully filled variant - a simple word replacement on the target side works well. However, simple masking also produces ``look \texttt{\_} left \texttt{\_}" which produces widely different targets for different values of \texttt{\_} and \texttt{\_}. Consider, 
    
    look \emph{around} left  $\rightarrow$ \texttt{I\_TURN\_LEFT I\_LOOK I\_TURN\_LEFT I\_LOOK I\_TURN\_LEFT I\_LOOK I\_TURN\_LEFT I\_LOOK} \newline \newline
    look \emph{opposite} left $\rightarrow$ \texttt{I\_TURN\_LEFT I\_TURN\_LEFT I\_LOOK}

  To discourage picking candidates like the latter one, we compute an \emph{alignment distance} between the candidate and its equivalent on the target side. 
For each candidate \(P\), gather the set of source sentences
\(\mathcal{S}(P)=\{s_1,s_2,\ldots,s_n\}\) in which the pattern occurs,
with each source sentence \(s_i\) paired to a target sentence \(t_i\).
For every \(s_i \in \mathcal{S}(P)\), calculate its Levenshtein
(edit) distance to every other \(s_j\;(j\neq i)\) in the same set and
select the \emph{nearest neighbour},
\(\mathrm{NN}(s_i)\)—the source sentence that minimises this distance.
Let \((t_i,t_j)\) denote the target sentences aligned with
\((s_i,\mathrm{NN}(s_i))\).

Define
\[
\Delta_i \;=\;
\left|
  \mathrm{len}(t_i)\;-\;\mathrm{len}(t_j)
\right|
\]
as the absolute difference in their word counts.
The resulting \emph{misalignment score} ($MS$) for pattern \(P\) is the
average of these differences:
\[
\operatorname{MS}(P)
\;=\;
\frac{1}{|\mathcal{S}(P)|}\,
\sum_{i=1}^{|\mathcal{S}(P)|}\Delta_i .
\]

A lower misalignment score indicates that source sequences are more aligned to the target sequences. A pseudo-construction has a low misalignment score when swapping different words into its slots still produces target sentences that look much the same. We add this score as a penalty to the beam search to pick candidates which are more aligned. 

\end{itemize}

Once the source files (train and test) are segmented, we prepare the data for the next stage. For every sentence in the source files, we replace the underscores with slot tokens such as \texttt{W\_n} where \texttt{n} refers to the slot number. We save the mapping between the slot tokens and the original words.

The SCAN data consists singleton rules such as \emph{jump} $\rightarrow$ \texttt{I\_JUMP}. We treat this as a bidirectional lexicon. Whenever a token in a target sentence appears in the lexicon, we lookup the source word and then replace it with the associated slot token. For example: 

\subsection{Training}

We use the sequence to sequence transformer architecture as the base model for training purposes, and use the JoeyNMT toolkit \citep{kreutzer-etal-2019-joey} to train all the models. The model architecture has an encoder and a decoder each with 4 layers and 4 attention heads with embedding size of 256 and the feed forward layer with the size of 1024. The models are trained for 30 epochs using the NOAM scheduler \citep{vaswani2017attention}. Prior to evaluation, we swap back the slot tokens predicted sequence through the mapping saved earlier.

\section{Results}

The performance on both the splits (ADD JUMP, AROUND RIGHT) is significantly better than the baseline transformer (\ref{tab:data-size}) which indicates that we have succeeded in encoding a degree of generalisation through the pseudo-constructions. Overall, they capture reusable structure absent from the flat surface strings, enabling the model to generalise compositionally.

\section{Data Efficiency}

Compositionality theory posits that exploiting compositional structure enables grasping abstract patterns from far fewer training examples than treating data only at the surface level (\citet{chomsky1957syntactic}, \citet{chomsky1965aspects}, \citet{fodor1988connectionism}). We test this by training models with smaller samples of the SCAN splits.

After segmentation of the training source file, each sentence is transformed into a series of pseudo-constructions in such a way that multiple sentences might fall into the same resultant sentence type. 
\\
look opposite left twice and walk twice $\rightarrow$ \texttt{ ( W\_1 opposite W\_2 twice ) and (W\_3 thrice)} \\
jump opposite right twice and run twice $\rightarrow$ \texttt{ ( W\_1 opposite W\_2 twice ) and (W\_3 thrice)} 

To assess the data efficiency of our method we constructed \emph{sentence type–balanced training subsets}, retaining every sentence type but varying the per-type quota $k\in\{1,3,5,10,25\}$.  This produces monotonic subsamples ranging from 5 \% to 60 \% of the original corpus while guaranteeing full coverage. (Table \ref{tab:data-size}).

On the ADD JUMP split, with only \emph{$k=10$} examples per type—approximately \emph{39 \%} of the full training data—the model attains \emph{40.7 \%} accuracy, not far from the \emph{47.8 \%} trained on entire set. 

For the AROUND RIGHT at the same \emph{$k=10$} mark the model reaches merely \emph{9.4 \%}, less than half of the \emph{20.4 \%} full-data accuracy, and increasing to \emph{$k=25$} (≈ 60 \%) yields only a marginal gain to \emph{10.2 \%}.  This pronounced gap reflects the split’s higher compositional complexity: mastering the nested “around DIR” construction with repetition operators may require substantially more evidence than the shallow “add jump” pattern.

The pseudo-construction bias confers strong sample-efficiency benefits on syntactically simple splits (ADD JUMP), but this may not scale to harder generalisation problems (AROUND RIGHT).

\begin{table}[t]
\centering
\small
\begin{tabular}{llrrr}
\hline
\textbf{Split} & \textbf{$k$ / type} & \textbf{Acc.\ (\%)} & \textbf{Size} & \textbf{\%} \\
\hline
Around Right  & 1    &  2.77 &   741 &   5 \\
             & 3    &  5.98 &  2,042 &  13 \\
             & 5    &  8.66 &  3,166 &  21 \\
             & 10   &  9.38 &  5,423 &  36 \\
             & 25   & 10.18 &  9,175 &  60 \\ 
             & Full    & 20.73 &  15,225 &   100 \\
\hline \\
Add Jump    & 1    & 12.79 &   666 &   5 \\
             & 3    & 20.50 &  1,972 &  13 \\
             & 5    & 29.44 &  3,178 &  22 \\
             & 10   & 40.69 &  5,660 &  39 \\
             & Full    & 47.81 & 14,670 &   100 \\
\hline
\end{tabular}
\caption{\label{tab:data-size}
Accuracy as the training set is reduced to \(k\) examples per type.}
\end{table}

\begin{figure}[t]
  \centering
  \includegraphics[width=\columnwidth]{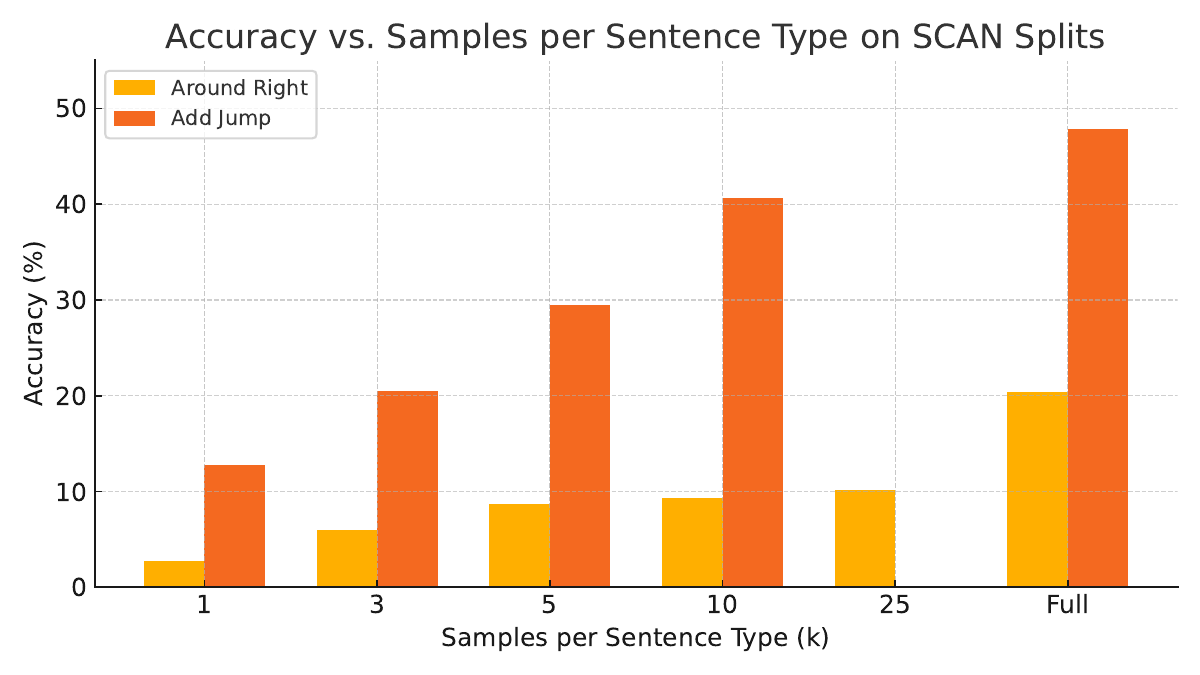}
  \caption{Accuracy versus percentage of full training data for the
           \textsc{Around~Right} and \textsc{Add~Jump} SCAN splits.}
  \label{fig:data-size-curve}
\end{figure}

\section{Conclusion}

While we define pseudo-constructions operationally as automatically mined templates, they can be seen as computational approximations to Construction Grammar’s notion of conventionalised form–meaning pairings. Unlike fully fossilised or community-shared constructions, pseudo-constructions are data-driven and context-specific, yet they capture structural regularities that support compositional generalisation. Thus, while our primary aim is methodological, the results also lend indirect support to the constructionist hypothesis that access to reusable schematic patterns is crucial for systematic generalisation. We leave a fuller exploration of their linguistic plausibility and theoretical integration to future work.

A deeper look into errors showed us that our method for finding pseudo-constructions can make several mistakes. For instance, while at first sight ``turn around right" and ``walk around right" seem to follow the same pattern, their corresponding outputs can vary significantly - this can lead to confusion and failure if the word ``turn" is masked away. 

We call for more robust approaches into finding constructions in text and for future work into deeper integration of construction processing into neural models.

\bibliographystyle{acl_natbib}
\bibliography{anthology,acl2021}


\end{document}